\documentclass[letterpaper]{article} 
\usepackage{aaai2026}  
\nocopyright
\usepackage{times}  
\usepackage{helvet}  
\usepackage{courier}  
\usepackage[hyphens]{url}  
\usepackage{graphicx} 
\usepackage{natbib}  
\usepackage{caption} 
\usepackage{algorithm}

\usepackage{booktabs}       
\usepackage{amsfonts}       
\usepackage{nicefrac}       
\usepackage{microtype}      

\usepackage{multirow}
\usepackage{algpseudocode}
\usepackage{amsmath}
\usepackage{amssymb} 

\usepackage{newfloat}
\usepackage{listings}
\DeclareCaptionStyle{ruled}{labelfont=normalfont,labelsep=colon,strut=off} 
\floatstyle{ruled}
\newfloat{listing}{tb}{lst}{}
\floatname{listing}{Listing}
\title{StableGS: A Floater-Free Framework for 3D Gaussian Splatting}
\author{
    Luchao Wang,
    Qian Ren,
    Kaimin Liao,
    Hua Wang,
    Zhi Chen,
    Yaohua Tang
}
\affiliations{
    Moore Threads AI

    tangyaohua28@gmail.com
}

\usepackage{bibentry}

\begin{document}

\maketitle

\begin{abstract}
3D Gaussian Splatting (3DGS) reconstructions are plagued by stubborn ``floater" artifacts that degrade their geometric and visual fidelity. We are the first to reveal the root cause: a fundamental conflict in the 3DGS optimization process where the opacity gradients of floaters vanish when their blended color reaches a pseudo-equilibrium of canceling errors against the background, trapping them in a spurious local minimum. To resolve this, we propose StableGS, a novel framework that decouples geometric regularization from final appearance rendering. Its core is a Dual Opacity architecture that creates two separate rendering paths: a ``Geometric Regularization Path" to bear strong depth-based constraints for structural correctness, and an ``Appearance Refinement Path" to generate high-fidelity details upon this stable foundation. We complement this with a synergistic set of geometric constraints: a self-supervised depth consistency loss and an external geometric prior enabled by our efficient global scale optimization algorithm. Experiments on multiple benchmarks show StableGS not only eliminates floaters but also resolves the common blur-artifact trade-off, achieving state-of-the-art geometric accuracy and visual quality.
\end{abstract}

\section{Introduction}
\label{sec:intro}

Synthesizing novel photorealistic views from multi-view images is a core topic in computer graphics and vision. 3DGS \cite{kerbl20233d} marks a significant breakthrough, using an explicit Gaussian-based representation to achieve real-time rendering at top-tier quality and greatly advancing Novel View Synthesis (NVS). Despite its success, 3DGS reconstructions are plagued by stubborn artifacts known as ``floaters"-erroneous, semi-transparent Gaussian clusters that degrade geometric accuracy and visual realism. Existing methods attempt to mitigate floaters but often lack a deep analysis of their root cause, thus failing to offer a targeted solution.

Our in-depth analysis reveals the root cause of floaters for the first time: an inherent conflict between color and geometry optimization in the 3DGS pipeline. We find that floaters become trapped in a spurious local minimum when their alpha-blended color reaches a pseudo-equilibrium with the background across multiple views. This state of canceling color errors causes the opacity gradient to vanish, making the floaters impossible to remove via standard optimization and exposing a fundamental limitation of the framework.

To break this optimization deadlock, we propose StableGS. While introducing geometric constraints is a natural solution, directly applying them to the rendered output harms appearance details. Our core contribution, the Dual Opacity GS model, resolves this trade-off by decoupling geometric regularization from final appearance rendering. It uses an auxiliary opacity parameter to create two distinct rendering paths: a `Geometric Regularization Path' for receiving harsh geometric constraints, and a `Appearance Refinement Path' for high-fidelity appearance. This design isolates the negative impacts of regularization, enabling a robust strategy built upon two complementary depth-based losses. First, a Self-supervised Depth Consistency Loss enforces internal geometric integrity without external data by penalizing inconsistencies between depth maps rendered from different views. Second, to handle challenging weakly-textured regions, we introduce a geometric prior from monocular depth estimation, made viable for large-scale scenes by our novel and efficient global scale optimization algorithm that transforms the intractable scale alignment problem into a compact one.

In summary, our main contributions are:
\begin{itemize}
    \item We are the first to identify the root cause of floaters as a gradient vanishing problem and propose a Dual Opacity architecture to decouple geometry from appearance. This design enables a robust regularization scheme-combining self-supervised consistency and a globally-optimized prior-to effectively eliminate floaters without harming visual quality.
    \item Extensive experiments demonstrate that StableGS not only eliminates floaters but also resolves the common blur-artifact trade-off, achieving superior performance on challenging benchmarks.
\end{itemize}

\section{Relate Work}
\label{sec:relatework}

\noindent\textbf{Novel View Synthesis} NVS aims to infer unseen perspectives from monocular image collections. NeRF~\citep{mildenhall2021nerf} pioneered MLP-based radiance field modeling with volume rendering, inspired extensive follow-up research. Subsequent works like NeRF++~\citep{zhang2020nerf++}, Mip-NeRF~\citep{barron2021mip}, and Mip-NeRF360~\citep{barron2022mip} enhanced rendering quality, where Mip-NeRF360 achieved anti-aliasing through ray-cone encoding and suppressed floaters via distortion loss. Meanwhile, DVGO~\citep{sun2022direct}, Instant-NGP~\citep{muller2022instant}, and TensoRF~\citep{chen2022tensorf} accelerated rendering to near real-time speeds. The emergence of 3DGS~\citep{kerbl20233d} established a new paradigm, employing explicit 3D Gaussian primitives to match Mip-NeRF360's quality while significantly surpassing its efficiency, making it the current mainstream approach.

\noindent\textbf{3DGS Improvements} 3DGS~\citep{kerbl20233d} revolutionized NVS through rasterized Gaussian rendering with exceptional performance. Mip-Splatting~\citep{yu2024mip} reduced aliasing artifacts via low-pass filtering, while densification strategy improvements emerged through absGS~\citep{ye2024absgs} (gradient absolute sum) and GOF~\citep{yu2024gaussian} (combined gradient metrics). Geometric regularization approaches include 2DGS~\citep{huang20242d} projecting Gaussians to 2D surfaces and RaDeGS~\citep{zhang2024rade} employing normal map constraints. Though some works~\citep{huang20242d,zhang2024rade} adopt NeRF-style distortion loss for floater reduction, we demonstrate this approach inadequately addresses 3DGS-specific floaters by erroneously increasing outlier opacity. Learnable Signed Distance Field (SDF) based methods like GSDF~\citep{yu2024gsdf} and SuGaR~\citep{guedon2024sugar} mitigated floaters but suffered from resolution-dependent parameterization limiting scalability. The floater issue introduces color block artifacts into rendered images, degrading visual quality. However, no prior work has systematically investigated the root causes of this phenomenon, nor proposed effective solutions.

\noindent\textbf{Dense Stereo Models} Recent advancements in monocular depth estimation, such as DUSt3R~\citep{wang2024DUSt3R} and its successors like MASt3R~\citep{leroy2024grounding} and Splatt3R~\citep{smart2024splatt3r}, have provided powerful geometric priors. We select DUSt3R for our work due to its superior accuracy. Concurrently, depth regularization has become a common strategy to enhance Gaussian splatting. It has been employed to address data sparsity in few-shot NVS~\citep{chung2024depth} and large-scale captures~\citep{kerbl2024hierarchical}. Other methods utilize depth priors to directly improve geometry, for instance, by anchoring splats to prevent drift in robotic datasets~\citep{lee2024mode} or by developing unbiased depth rendering techniques for better multi-view consistency~\citep{chen2024pgsr}.

\section{Methodology}
\label{sec:methodology}


\subsection{Preliminaries}  

3DGS represents scenes using learnable 3D Gaussian ellipsoids $\mathcal{G} = \{G_k\}_{k=1}^K$, where each Gaussian $G_k$ is parameterized by its center $\mu_k \in \mathbb{R}^3$, opacity $\alpha_k \in [0,1]$, color $c_k \in \mathbb{R}^3$, scale $s_k \in \mathbb{R}^3$, and rotation quaternion $q_k \in \mathbb{R}^4$.  The opacity of $G_k$ at spatial point $x \in \mathbb{R}^3$ is given by:  
\begin{equation}  
\scriptsize
\sigma_k(x) = \alpha_k \exp\left(-\frac{1}{2}(x-\mu_k)^\top \Sigma_k^{-1}(x-\mu_k)\right)  
\label{equ:sigmakx}
\end{equation}  
where $\Sigma_k$ is the covariance matrix derived from the rotation and scale parameters. For simplicity, we use the color $c_k$ directly, rather than Spherical Harmonics.



Novel views are rendered using a differentiable rasterization pipeline. For each camera ray $r$, the $K_r$ Gaussians that intersect its corresponding pixel are sorted by depth. The final color $\bar{c}(r)$ and depth $\bar{d}(r)$ are then accumulated via alpha blending: $\bar{c}(r) = \sum_{k=1}^{K_r} c_k \sigma(x_k^r)\tau_k$ and $\bar{d}(r) = \sum_{k=1}^{K_r} d_k^r \sigma(x_k^r)\tau_k$, where $\sigma(x_k^r)$ is the opacity contribution of the $k$-th Gaussian along the ray, $d_k^r$ is its depth, and $\tau_k = \prod_{j=1}^{k-1} (1 - \sigma(x_j^r))$ is the transmittance. We use $\bar{(\cdot)}$ to denote rendered outputs. The model is optimized by minimizing a loss between the rendered image $\bar{C}$ and the ground-truth $C^\text{gt}$:
\begin{equation}
\small
\mathcal{L}_\text{3DGS} = (1-\lambda)\mathcal{L}_1(\bar{C}, C^\text{gt}) + \lambda \mathcal{L}_\text{D-SSIM}(\bar{C}, C^\text{gt})
\label{equ:loss3dgs}
\end{equation}
where $\mathcal{L}_1$ and $\mathcal{L}_\text{D-SSIM}$ are the L1 and D-SSIM losses.

\subsection{Floater Analysis in Gaussian Splatting}

Our in-depth research reveals that the persistence of floaters in well-trained models stems from a critical flaw in the optimization process: gradient vanishing during backpropagation. This makes it impossible for the model to fully eliminate geometric errors using only the standard color-based loss.

To clearly elucidate this mechanism, we analyze a simplified scenario focused on optimizing a floater cluster, which is traversed by $M$ rays with corresponding ground-truth colors $\{c_m^\text{gt}\}_{m=1}^M$. The objective is to minimize the L1 color error. To simplify the problem, we assume that the rest of the scene has been perfectly reconstructed, and that the background color observed by a ray after passing through the cluster is identical to the ground truth. While the theoretical optimum is for all floater opacities $\alpha$ to be zero, the optimization often gets trapped in a local minimum. In this state, the accumulated color $c_f$ from the alpha-blended floater cluster exhibits a key characteristic: it reaches a ``pseudo-equilibrium" where the signs of the color differences against the ground truth nearly cancel each other out across all $M$ views: $\sum_{m=1}^M \text{sign}(c_f - c_m^{\text{gt}}) \approx 0$.
For the L1 loss, this condition directly causes the gradient of the accumulated color $c_f$ to approach zero. According to the chain rule, this in turn triggers a cascading vanishing of the gradients for both the opacity $\alpha$ and color $c$ parameters. We have verified the generality of this phenomenon across various loss functions, including L1 and MSE.

This gradient vanishing phenomenon prevents the optimizer from effectively reducing the opacity of the floater Gaussians, making them impossible to ``remove" geometrically. The model becomes trapped in a local minimum that is geometrically incorrect but has a very low color loss, revealing the inherent limitation of relying solely on color-based loss to optimize geometry. However, the error of floaters in the geometric dimension-namely, their depth-is significant and unambiguous. This inspires us to introduce depth-based loss functions, which can provide an independent and stronger geometric supervision signal to potentially pull the optimization out of these spurious local minima.

\subsection{Dual Opacity GS}

While introducing a depth-based loss is a viable path to resolving the gradient vanishing problem, its direct application within the standard 3DGS framework leads to a loss of rendering detail. The strong gradients from a depth loss, while effective at correcting geometric errors like floaters, are derived from supervision signals (either multi-view constraints or external priors) that inevitably contain noise and inaccuracies. These imperfect yet powerful constraints can pollute all Gaussian parameters, resulting in smoothed details and blurred textures. Moreover, for translucent objects such as glass or water, direct depth supervision can severely harm rendering precision.

To break this dilemma, we propose a novel architecture: the \textbf{Dual Opacity Gaussian Splatting} model. The core purpose of this design is not to solve the gradient vanishing problem directly, but rather to establish a ``firewall" mechanism that can effectively bear depth-based regularization while isolating its adverse effects from the final appearance rendering. Specifically, we introduce an additional learnable auxiliary opacity parameter, $\alpha'_k \in [0,1]$, for each Gaussian $G_k$. The original opacity parameter, $\alpha_k$, now serves as the representation of the base geometry, while $\alpha'_k$ is used to modulate the final visual transparency. Based on this, we define two distinct opacity functions:
\begin{itemize}
    \item \textbf{Geometric Opacity}: $\sigma_k(x)$, identical to the standard 3DGS definition (Eq. \ref{equ:sigmakx}), using only $\alpha_k$.
    \item \textbf{Rendering Opacity}: $\sigma'_k(x) = \alpha'_k \sigma_k(x)$, which is determined by both opacity parameters.
\end{itemize}

Correspondingly, our model possesses two functionally decoupled rendering paths:
\begin{itemize}
    \item \textbf{Geometric Regularization Path}: Renders an intermediate color map $\bar{C}^s$ and depth map $\bar{D}^s$ using the geometric opacity $\sigma(x)$. The output of this path serves exclusively as the supervision target for our subsequent depth-based loss functions.
    \item  \textbf{Appearance Refinement Path}: Renders the final, high-quality color map $\bar{C}^o$ and depth map $\bar{D}^o$ using the rendering opacity $\sigma'(x)$.
\end{itemize}

This decoupling architecture is the key to making depth-based losses truly effective. We can confidently apply powerful depth constraints on the output of the geometric regularization path, $\bar{D}^s$. The resulting strong gradients act directly on the base geometric parameters, effectively eliminating floaters and correcting the overall geometry. Any resulting detail loss or appearance distortion is confined within this geometric path. The Appearance Refinement Path, thanks to the learnable auxiliary parameter $\alpha'_k$, is shielded from these coarse geometric gradients. It builds upon the ``geometric scaffold", whose structure has already been ensured by the geometric path, and focuses on learning the fine-grained color and translucency effects needed to achieve a high-quality match with the ground-truth image $C^\text{gt}$.

The model's overall loss function consists of a color loss $\mathcal{L}_{\text{color}}$ and a geometry loss $\mathcal{L}_{\text{geometry}}$:
\begin{equation}
\small
    \mathcal{L}_\text{total} = \mathcal{L}_\text{color} + \lambda_\text{geo} \mathcal{L}_\text{geometry}
\end{equation}
The color loss $\mathcal{L}_{\text{color}}$ includes supervision for both paths $\mathcal{L}_\text{color} = \mathcal{L}_\text{3DGS}(\bar{C}^o, C^\text{gt}) + \lambda_s \mathcal{L}_1(\bar{C}^s, C^\text{gt})$. 
The first term is the primary appearance loss applied to the final rendered output. The second term, $\mathcal{L}_1(\bar{C}^s, C^\text{gt})$, is an auxiliary loss weighted by $\lambda_s$ that encourages the base geometric representation (dominated by $\alpha_k$) to explain the opaque parts of the scene as well as possible. Consequently, the gradients for the auxiliary parameter $\alpha'_k$ will primarily come from regions that the geometric path cannot perfectly model, such as translucent surfaces and fine structures. The $\mathcal{L}_{\text{geometry}}$ term comprises all regularization terms applied to the geometric depth map $\bar{D}^s$, which we detail in the next section.

\subsection{Self-supervised Depth Consistency}

\begin{figure}[t]
  \centering
    \includegraphics[width=1.0\linewidth]{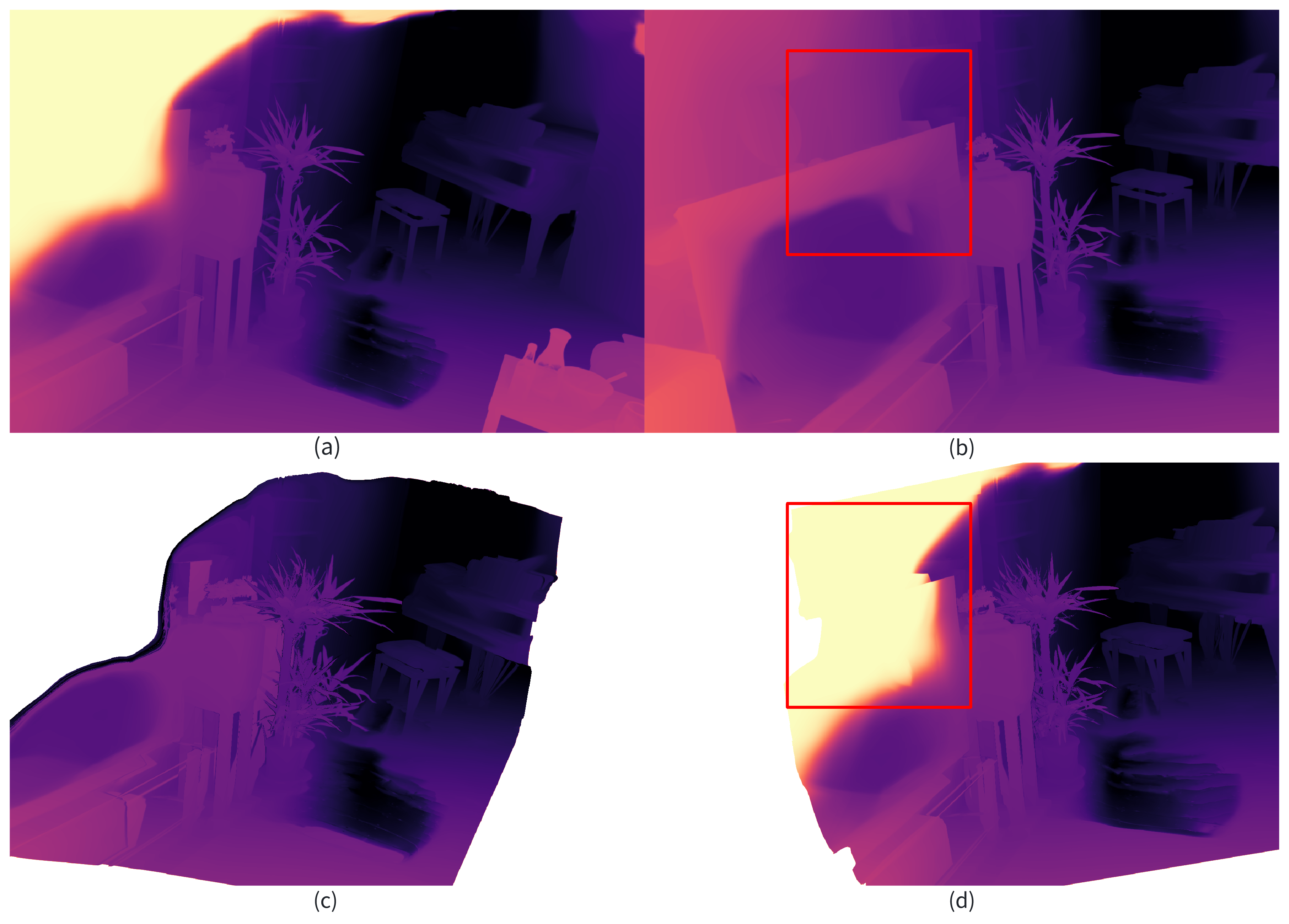}
    \caption{The figures show RadeGS rendering results of two example images from the room dataset training set (Fig. (a) and (b)) and their warped outcomes (Fig. (c) and (d)). A floater visible in the upper-left corner of Fig. (a) is projected into the red-boxed region of Fig. (d), while the corresponding red-boxed region in Fig. (b) displays accurate depth measurements. By enforcing constraints from Fig. (b) onto Fig. (d), the floater can be effectively eliminated.}
    \label{fig:touying}
\end{figure}

Building on our decoupled architecture, we first introduce a self-supervised geometric constraint. It leverages the scene's inherent multi-view consistency to correct flawed geometric structures, particularly floaters, without relying on any external data. The core idea is that a depth map rendered from view $i$, $\bar{D}_i^s$, when warped to another view $j$, should match the depth map rendered directly from that view, $\bar{D}_j^s$. Floaters, with their erroneous depth values (typically an average of their own depth and the background's), violate this principle. Warping these incorrect depth values to another view creates a large discrepancy against the correct geometry, providing a strong gradient signal for their elimination, as shown in Fig. \ref{fig:touying}. Specifically, we define a depth consistency loss, $\mathcal{L}_{\text{consis}}$, which is applied exclusively to the depth maps $\bar{D}^s$ from our Geometric Regularization Path. For a set of selected image pairs $\mathcal{P}$, the loss is:
\begin{equation}
    \scriptsize
    \mathcal{L}_{\text{consis}} = \sum_{(i,j)\in \mathcal{P}} \left( \| \mathcal{W}_{i\to j}(\bar{D}_i^{s}) - \bar{D}_j^{s} \|_1 + \| \mathcal{W}_{j\to i}(\bar{D}_j^{s}) - \bar{D}_i^{s} \|_1 \right)
\end{equation}
Here, $\mathcal{W}_{i\to j}$ is the differentiable warping operator, which uses known camera poses to reproject points from view $i$ to $j$. The loss is computed only over validly projected regions. To ensure the stability and effectiveness of this loss, the training pairs in $\mathcal{P}$ are selected using three criteria: 1) The number of co-visible feature points from COLMAP between two views must be greater than $30$ to ensure sufficient overlap; 2) The relative rotation angle is constrained to $16^\circ-60^\circ$ to prevent extreme baselines; 3) The percentage of feature points conforming to a homography matrix must be below 0.8 to suppress degenerate cases arising from near-pure rotation.


\subsection{External Priors with Efficient Global Optimization}

To address the decline in geometric accuracy in weakly-textured or texture-less regions (e.g., white walls, smooth floors) due to insufficient visual features, we introduce external geometric priors from the monocular depth estimation model DUSt3R. For any image pair $(i, j)$, DUSt3R outputs two associated point maps, $X^{1,1}_{i,j}$ and $X^{2,1}_{i,j}$, along with their confidence maps, $C^{1,1}_{i,j}$ and $C^{2,1}_{i,j}$. Here, $X^{1,1}_{i,j}$ represents the 3D coordinates of points from image $i$ in its own camera frame, while $X^{2,1}_{i,j}$ represents points from image $j$ in image $i$'s camera frame. The critical bottleneck, however, is that each pairwise prediction exists in an independent, unknown scale, making it unusable for global 3D reconstruction.

To obtain a set of depth priors with a consistent global scale, we propose a novel and efficient global scale optimization algorithm to compute the optimal scale factors $S = \{s_{i,j}, s_{j,i}\}_{(i,j)\in\mathcal{P}}$ for all image pairs. Our optimization objective consists of two core components.

\subsubsection{Pairwise Point Cloud Alignment} 
This objective requires that for each pair $(i, j)$, their corresponding point clouds align in 3D space after being transformed by their respective scale factors and camera poses, $T^i$ and $T^j$. The alignment error is given by \scriptsize $
\mathcal{L}_{\text{pair}}(i,j) = 
\left\|  [\mathcal{C}(i,j),\mathcal{C}(j,i)]  \odot ([s_{i,j}T^i,-s_{j,i}T^j] \left[ \begin{matrix}X^{1,1}_{i,j} & X^{2,1}_{i,j} \\ X^{2,1} _{j,i}&X^{1,1}_{j,i}\\ \end{matrix} \right] \right\|$\normalsize, where $\odot$ denotes element-wise multiplication and \scriptsize  $\mathcal{C}(i,j)=\frac{C^{1,1}_{i,j}C^{2,1}_{j,i}}{C^{1,1}_{i,j}+C^{2,1}_{j,i}}$\normalsize. Minimizing this alignment error on the full, raw point clouds is computationally intractable due to immense memory and processing overhead.

Our core contribution is a matrix-factorization-based equivalent transformation that rigorously simplifies this high-dimensional point cloud alignment problem into a low-dimensional optimization over just 7 ``equivalent points".  Consequently, we can efficiently express the pairwise alignment loss as \scriptsize
$\mathcal{L}_{\text{pair-eff}}(i,j) = \left\| w_{i,j} \odot (s_{i,j}T_ix_{i,j} - s_{j,i}T_jy_{i,j}) \right\|_2^2$ \normalsize, where $x_{i,j}, y_{i,j} \in \mathbb{R}^{3\times 7}$ and $w_{i,j} \in \mathbb{R}^7$ are the equivalent points and corresponding weights decomposed from the original point clouds.

\noindent \textit{
\textbf{Equivalence Proof}: For a general weighted point set matching problem formulated as \(||\mathcal{J}\odot(A_1Y_1^\top+b_1-A_2Y_2^\top-b_2)||\), where \(A_1,A_2 \in \mathbb{R}^{3\times 3}\) and \(b_1,b_2 \in \mathbb{R}^{3\times 1}\) represent arbitrary 3D affine transformations, \(Y_1,Y_2 \in \mathbb{R}^{N \times 3}\) denote source and target points, and \(\mathcal{J} \in \mathbb{R}^{1 \times N}\) denotes weights. Let \(b = b_1 - b_2\), the formulation can be equivalently expressed as:
\begin{equation}
\scriptsize
\begin{aligned}
& ||\mathcal{J} \odot ([A_1,-A_2,b] [Y_1,Y_2,1]^\top)|| \\
= & [A_1,-A_2,b] \underbrace{ [Y_1,Y_2,1]^\top diag(\mathcal{J}^2)[Y_1,Y_2,1]}[A_1,-A_2,b]^\top \\
= & [A_1,-A_2,b] V^\top diag(\xi^2)V [A_1,-A_2,b]^\top 
\end{aligned}
\nonumber
\end{equation}
where \(V^\top \text{diag}(\xi^2)V\) constitutes the eigenvalue decomposition of the bracketed term, with \(V \in \mathbb{R}^{7 \times 7}\) and \(\xi \in \mathbb{R}^7\). This can be reformulated into a point-matching form \(||w \odot ([A_1,-A_2,b] [x,y,1]^\top)||\), where \(w = \xi V_{[:7]}\), \(x = V_{[:1-3]}/V_{[:7]}\), and \(y = V_{[:4-6]}/V_{[:7]}\), with \(V_{[:.]}\) denoting corresponding columns of \(V\).
}

\setlength{\tabcolsep}{4pt}
\begin{table*}[ht]
\centering
\small
\begin{tabular}{lcccccccccccc}
\toprule
\multirow{2}{*}{\footnotesize{Architecture}} & \multicolumn{3}{c}{mipnerf360 indoor} & \multicolumn{3}{c}{mipnerf360 outdoor} & \multicolumn{3}{c}{Blender} & \multicolumn{3}{c}{Tanks\&Temples} \\ \cline{2-13} 
 & PSNR$\uparrow$ & SSIM$\uparrow$ & LPIPS$\downarrow$ & PSNR$\uparrow$ & SSIM$\uparrow$ & LPIPS$\downarrow$ & PSNR$\uparrow$ & SSIM$\uparrow$ & LPIPS$\downarrow$ & PSNR$\uparrow$ & SSIM$\uparrow$ & LPIPS$\downarrow$ \\ \hline
3DGS(updated)                                       & 31.160 & 0.9255 & 0.1859 & 26.348 & 0.7882 & 0.2058 & 29.646 & 0.9070 & 0.2387 & 23.933 & 0.8562 & 0.1697 \\ \hline
Mip-Splatting                                       & 30.971 & 0.9242 & 0.1870 & 26.384 & 0.7912 & 0.2051 & 29.622 & 0.9050 & 0.2431 & 23.836 & 0.8531 & 0.1737 \\
GOF                                                 & 30.465 & 0.9178 & 0.1888 & 26.566 & 0.8052 & 0.1852 & 29.569 & 0.9059 & 0.2427 & 23.592 & 0.8530 & 0.1682 \\
RaDeGS                                              & 30.509 & 0.9213 & 0.1887 & 26.704 & \textbf{0.8119} & 0.1753 & 29.530 & 0.9073 & 0.2416 & 23.897 & 0.8560 & 0.1673 \\ 
Taming-3DGS                                         & 31.107 & 0.9225 & 0.1926 & 26.645 & 0.7997 & 0.1948 & 30.031 & 0.9065 & 0.2395 & 24.117 & 0.8546 & 0.1732 \\
\textbf{StableGS} w/o $\mathcal{L}_\text{prior}$    & \textbf{31.548} & \textbf{0.9283} & \textbf{0.1762} & \textbf{26.733} & 0.8102 & \textbf{0.1716} & \textbf{30.123} & \textbf{0.9111} & \textbf{0.2341} & \textbf{24.165} & \textbf{0.8577} & \textbf{0.1628} \\ \hline \hline
3DGS+depth                                          & 31.029 & 0.9245 & 0.1861 & 26.379 & 0.7875 & 0.2058 & 29.976 & 0.9086 & 0.2363 & 24.070 & 0.8558 & 0.1668 \\
\textbf{StableGS}                                   & 31.542 & 0.9281 & 0.1761 & 26.738 & 0.8105 & 0.1715 & 30.190 & 0.9113 & 0.2347 & 24.196 & 0.8590 & 0.1610        \\\bottomrule
\end{tabular}
\label{table:psnr}
\caption{Performance comparison on mipnerf360, Blender and Tanks\&Temples. \textbf{Bolded} indicates the best among methods without depth prior.}
\end{table*}

\begin{figure*}[ht]
  \centering
    \includegraphics[width=0.95\textwidth]{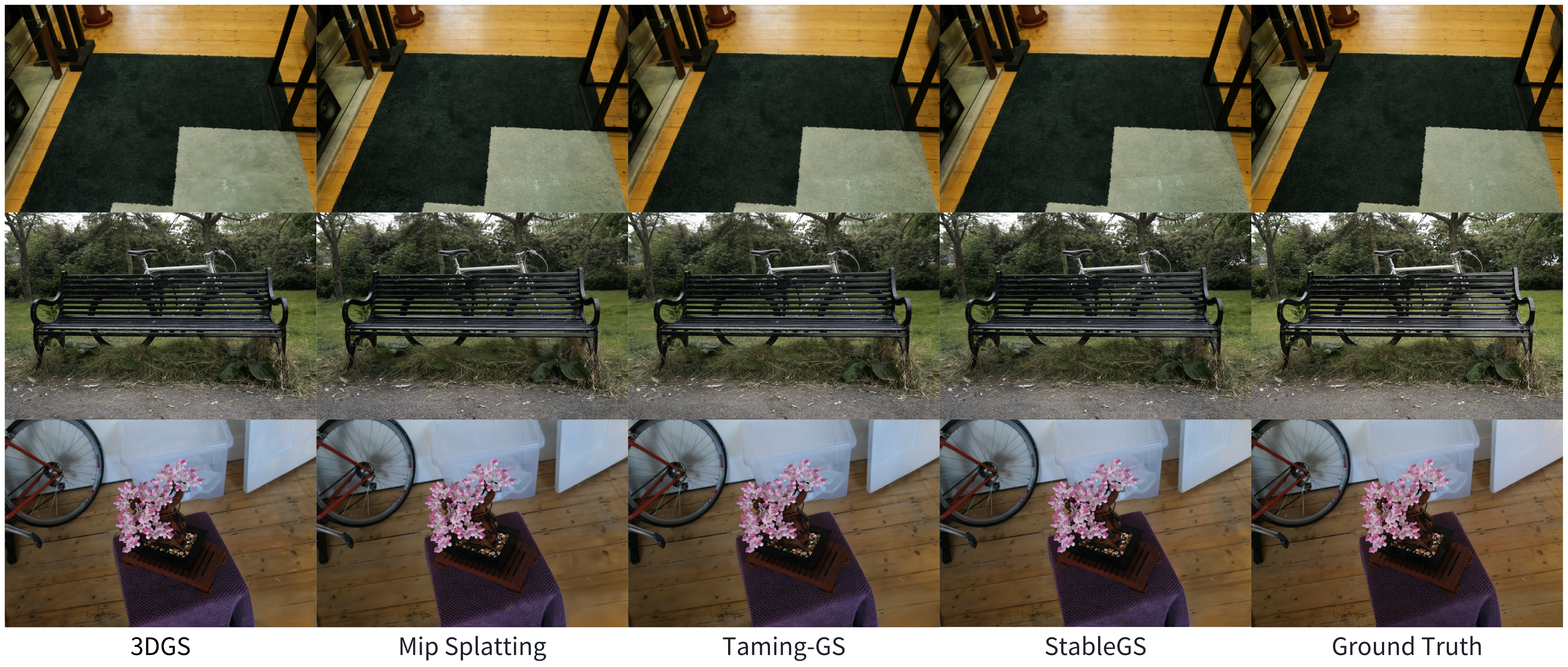}
    \caption{\textbf{blurry problem.}  Note the first row texture details of the carpet, the second row grass beneath the chair, and the third row floor details (particularly in the lower-right corner region). Our method preserves finer image details.}
    \label{fig:mohu}
\end{figure*}

\subsubsection{Cross-Pair Scale Consistency} To ensure a unified global scale, we further constrain the scale associated with the same image to be consistent across different pairs. We implement this with a regularization term, \scriptsize $\mathcal{L}_{\text{intra}}(i,j) = \left\|\mathcal{C}(i,a)\odot \left( s_{i,j}X^{1,1}_{i,j}-s_{i,a}X^{1,1}_{i,a}\right) \right\|_1$\normalsize, which penalizes the deviation of a pair's scale from that of a high-confidence ``anchor pair".  The anchor pair index $a = \arg\max_k (\sum C^{1,1}_{i,k})$ corresponds to the pairing with the highest sum of confidences for image $i$.

Combining these components, our final global scale optimization problem is formulated to efficiently solve for the scale factors $S$ using only the compact equivalent points:
\begin{equation}
\small
    \min_{S} \sum_{(i,j) \in \mathcal{P}} \left( \mathcal{L}_{\text{pair-eff}}(i,j) +  \mathcal{L}_{\text{intra}}(i,j)+\mathcal{L}_{\text{intra}}(j,i) \right) 
\end{equation}
In fact, we can also optimize the camera poses in the same manner to perform Structure-from-Motion (SfM) directly and achieve dense MVS reconstruction via TSDF.

After solving for the globally consistent scale factors $S$, we can generate a reliable set of depth priors, where each depth map $d_{i,j}=s_{i,j}X^{1,1}_{i,j}[z]$ is correctly scaled. We then define a geometric prior loss, $\mathcal{L}_\text{prior}$, which is applied to the depth maps $\bar{D}^s$ from the Geometric Regularization Path:
\begin{equation}
\small
\mathcal{L}_\text{prior} = \sum_{(i,j)\in \mathcal{P}} \left( \mathcal{D}(i,j) + \mathcal{D}(j,i) \right)
\end{equation}
where $\mathcal{D}(i,j)=\frac{1}{\|d_{i,j}\|_0 } \sum \left( C^{1,1}_{i,j}\odot |\bar{D^s_i}-d_{i,j}| \right)$, $\|\cdot\|_0$ is the L0 norm, and the summation is over all valid pixels $p$. By minimizing this loss, our model learns the correct geometry in weakly-textured regions, while our decoupled architecture ensures that noise from the priors does not pollute the final, high-quality appearance rendering. Finally, our total geometry loss is a weighted sum of the two regularization terms:
$$\mathcal{L}_\text{geometry} = \lambda_\text{consis}\mathcal{L}_\text{consis} + \lambda_\text{prior}\mathcal{L}_\text{prior}$$

\begin{figure*}[htb]
  \centering
    \includegraphics[width=0.95\textwidth]{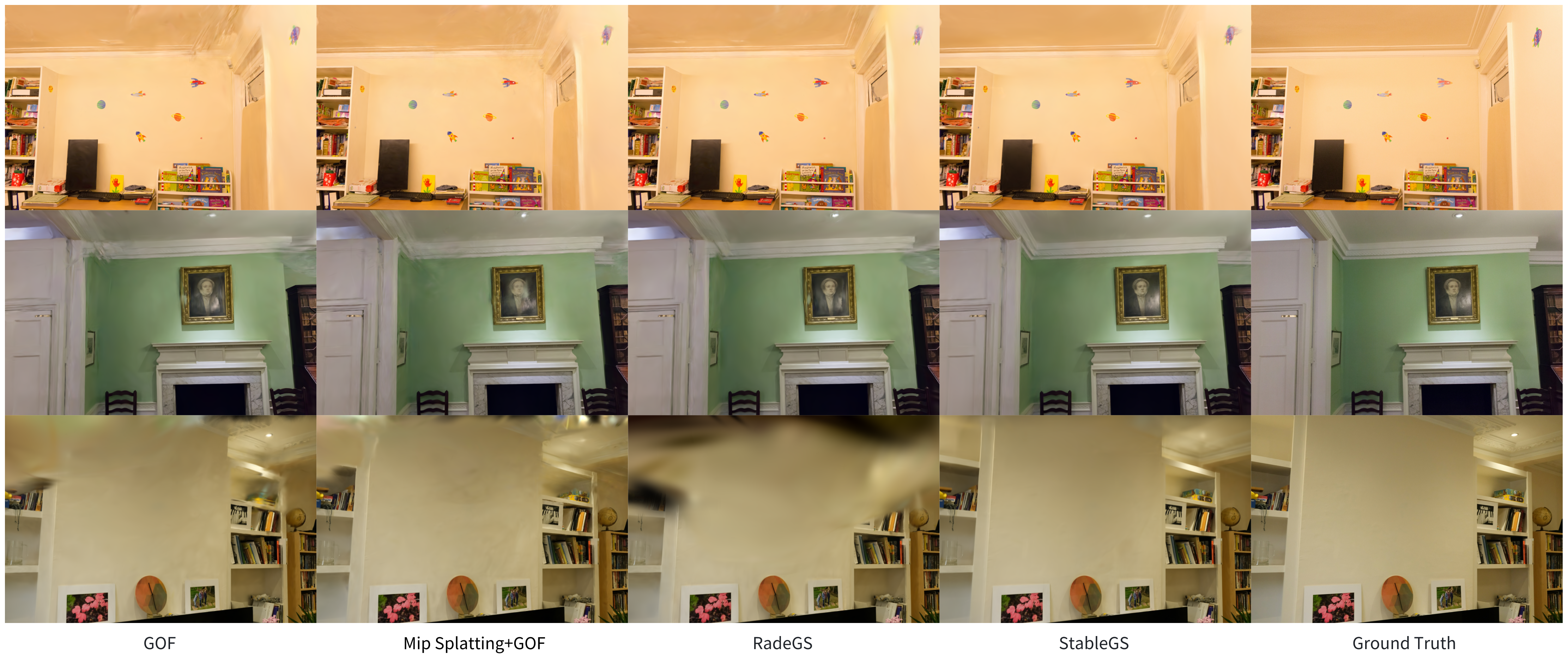}
    \caption{\textbf{floater problem.} Note the first row ceiling and upper-right wall region, the second row prominent artifacts in the upper-left and upper-right corners; and the third row wall surfaces. Our method significantly reduces artifacts and eliminates floater issues.}
    \label{fig:floater}
\end{figure*}

We name our complete framework \textbf{StableGS}.  The Dual Opacity GS model serves as the foundational architecture, providing decoupled paths for geometry and appearance. Upon this architecture, it coordinates two complementary regularization strategies: the self-supervised depth consistency loss enforces inherent structural integrity, while the globally optimized geometric prior specifically tackles challenging, texture-less regions. This systematic design enables StableGS to not only eliminate the floater artifacts of standard 3DGS but also to exhibit exceptional geometric accuracy and appearance fidelity when reconstructing challenging scenes.

\begin{figure*}[h!]
  \centering
    \includegraphics[width=0.95\linewidth]{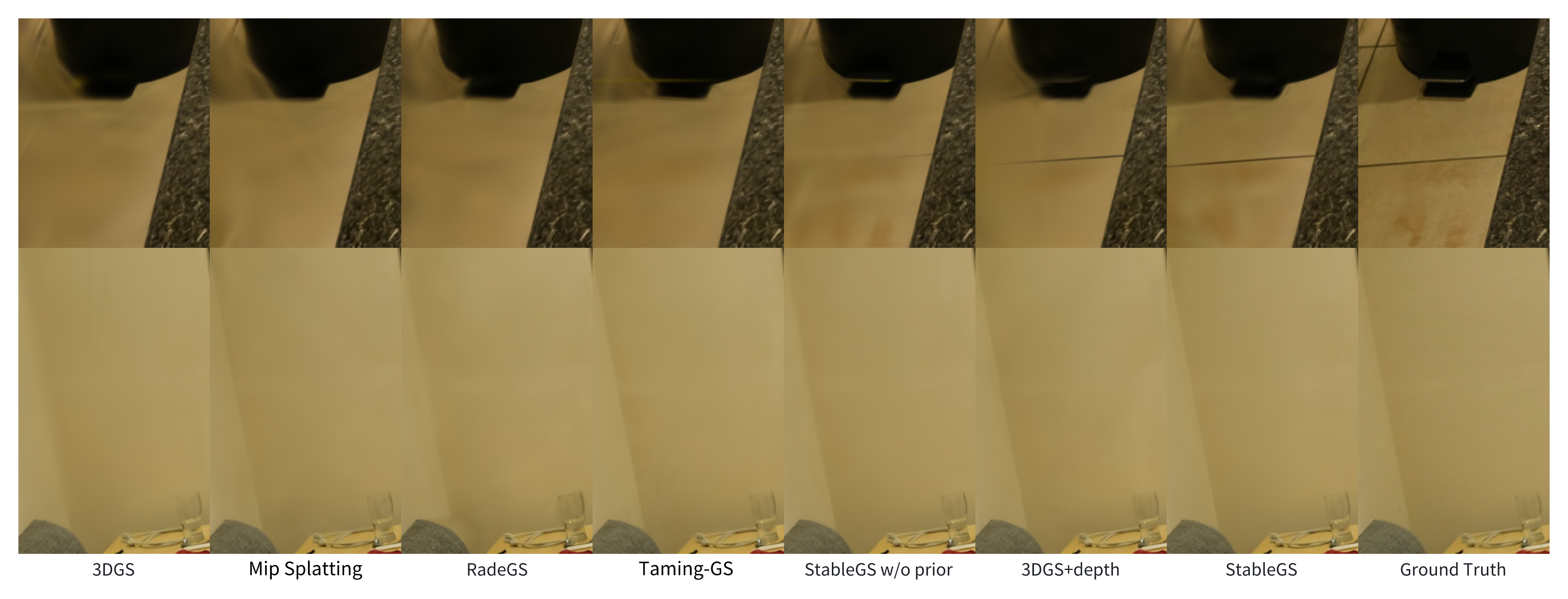}
    \caption{\textbf{Weakly-Textured Region Reconstruction.} Observe the tile gaps in the first row and wall corners in the second row. Only methods incorporating depth priors successfully restore these high-frequency features, with our approach achieving the sharpest reconstruction.}
    \label{fig:weakfeature}
\end{figure*}

\section{Experiments}
\label{sec:experiments}

\subsection{Experimental setting}
\label{sec: expsetting}

Our work implements differentiable depth rendering based on RaDeGS \cite{zhang2024rade} and utilizes the Taming-3DGS training pipeline. All experiments are run for 30,000 iterations on a single NVIDIA A100 GPU. We keep the learning rates and other hyperparameters consistent with the original 3DGS. For our proposed losses, we set the weights to $\lambda_{\text{consis}}=0.05$ and $\lambda_{\text{prior}}=0.005$ across all experiments.

We evaluate our method on three datasets: MipNeRF-360 \cite{barron2022mip}, Tanks \& Temples \cite{knapitsch2017tanks}, and Deep Blending \cite{hedman2018deep}. For the MipNeRF-360 dataset, we train on images at half resolution, with the longest edge scaled to 1600 pixels while preserving the aspect ratio. Consistent with prior work~\cite{kerbl20233d}, we use PSNR, SSIM, and LPIPS as image quality metrics. We compare our method against several strong baselines, including 3DGS~\citep{kerbl20233d} updated to the latest stable release from the official repository, Mip-Splatting~\citep{yu2024mip}, Taming-3DGS~\citep{mallick2024taming}, GOF~\citep{yu2024gaussian}, and RaDeGS. For a fair comparison, the final Gaussian counts for our StableGS and Taming-3DGS are controlled to be consistent with that of 3DGS.

\begin{figure*}[ht]
  \centering
    \includegraphics[width=0.95\textwidth]{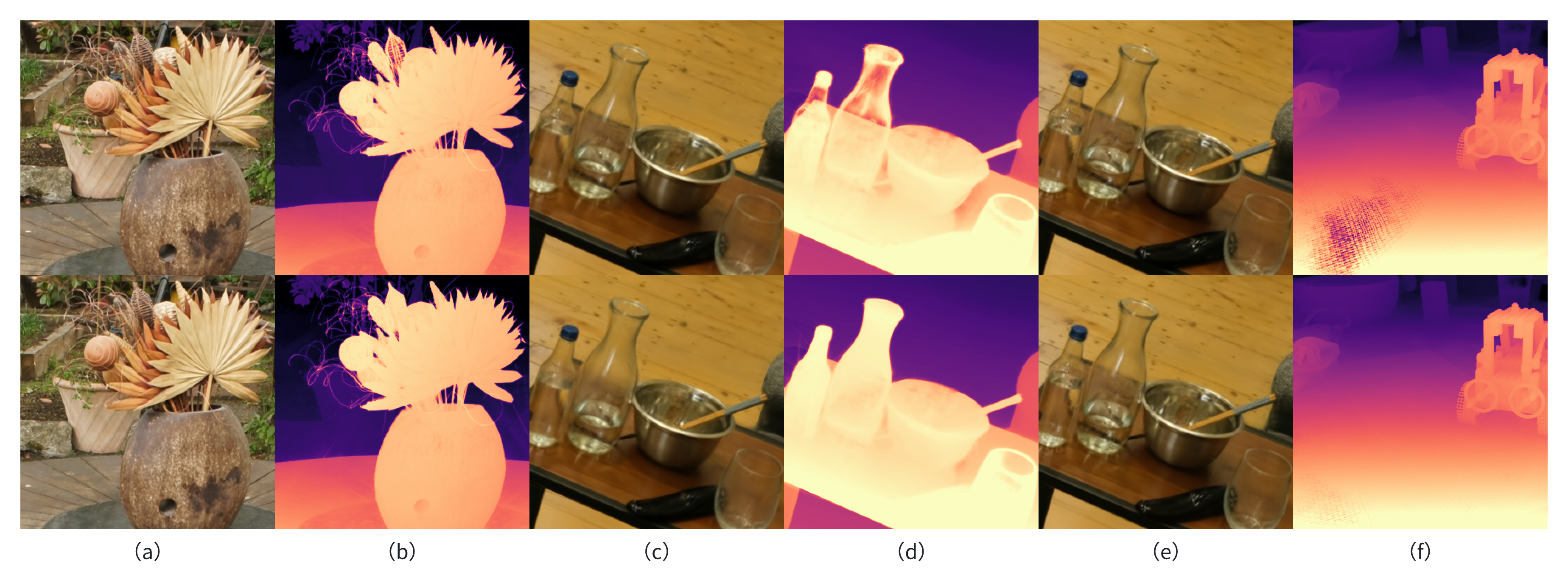}
    \caption{Fig. (a)-(d) compare the two rendering modes of Dual Opacity GS, with the first row depicting the Appearance Refinement Path and the second row the Geometric Regularization Path. Fig. (e) contrasts standard 3DGS (first row) and 3DGS with depth regularization(second row), demonstrating that direct depth integration incurs precision degradation. Fig. (f) presents depth map results from the kitchen dataset, comparing StableGS w/o $\mathcal{L}_\text{prior}$ (first row) and StableGS (second row), to validate the impact of depth priors on reconstruction fidelity.}
    \label{fig:depthloss}
\end{figure*}

\subsection{Comparison}
The results are presented in Table 1, showing that our method generally outperforms existing approaches. First, we compare our `StableGS w/o $\mathcal{L}_\text{prior}$' variant, which does not use a depth estimator, against the baselines. On the MipNeRF-360 indoor scenes, our method achieves substantial improvements over prior work, likely because the well-structured nature of these scenes allows our depth-based losses to be particularly effective. For outdoor scenes, both our method and RaDeGS show a significant lead, with StableGS performing slightly better. While RaDeGS's core innovation is constraining Gaussian normals, our method achieves high-quality results through its depth-based regularization. On the Blender and Tanks \& Temples datasets, our method also achieves the best results. Notably, our method achieves significantly better LPIPS scores across all experiments. We attribute this to the improvement in fine detail rendering that results from resolving the floater problem.

Furthermore, we compare against methods that use depth priors. Our full StableGS model significantly outperforms `3DGS with depth regularization' presented in \cite{kerbl2024hierarchical}, where the depth is estimated by DepthAnything \cite{depth_anything_v2}. We note that the metric improvements for both `3DGS with depth regularization' (over 3DGS) and our full `StableGS' (over `StableGS w/o $\mathcal{L}_\text{prior}$') are modest. This is because while depth priors can greatly improve reconstruction accuracy in weakly-featured regions, this improvement is not always fully reflected in standard image-space metrics. 

We will demonstrate this specific benefit in the subsequent qualitative analysis from three perspectives: resolving blurriness, eliminating floaters, and reconstructing weakly-textured regions. We will show through examples that all prior methods suffer from either blurriness or floaters, whereas StableGS simultaneously eliminates blur without introducing floaters. Moreover, by incorporating geometric priors, StableGS yields the best results in weakly-featured areas compared to existing methods.

\noindent\textbf{Blurriness}. As shown in Fig.~\ref{fig:mohu}, the original 3DGS exhibits noticeable blur. Mip-Splatting, despite solving the aliasing problem, exacerbates the blurriness. Taming-3DGS, despite its improved densification strategy, still suffers from blur. In contrast, our StableGS achieves the sharpest results.

\noindent\textbf{Floaters}. All baselines in Fig.~\ref{fig:floater} use the GOF densification strategy and they all exhibit floaters to varying degrees, while our result remains clean.

\noindent\textbf{Weakly-Textured Regions}. In Fig.~\ref{fig:weakfeature}, we compare the rendering performance of Mip-Splatting, Taming-3DGS, and RaDeGS in such regions. To demonstrate the specific contributions of our depth prior loss and dense initialization, we also show the result of our `StableGS w/o $\mathcal{L}_\text{prior}$' for comparison. We can see that StableGS achieves the best performance, producing the sharpest edges on challenging areas like wall corners.

\subsection{Ablation studies}
Our Dual Opacity GS architecture effectively isolates the adverse effects of depth-based losses. As shown in Fig.~\ref{fig:depthloss}-(b), inaccuracies in the depth prior cause the Geometric Regularization Path to render images with lost detail on the flower spikes. However, in the Appearance Rendering Path, these fine details are successfully recovered. Similarly, for translucent objects, which are inherently problematic for depth losses, Fig.~\ref{fig:depthloss}-(d) demonstrates that the Appearance Rendering Path correctly restores the semi-transparent regions. In contrast, Fig.~\ref{fig:depthloss}-(e) shows the result of applying a depth constraint directly to standard 3DGS on the same case; this leads to a significant loss of precision for the translucent bottle, blurring the background seen through it. Our full method, shown in Fig.~\ref{fig:depthloss}-(c), produces the correct, sharp rendering.

\noindent\textbf{Depth Consistency Loss}: We experimentally demonstrate the effect of the Depth Consistency Loss on translucent objects. As seen in Fig.~\ref{fig:depthloss}-(d), the Geometric Regularization Path treats the glass bottle as an opaque solid in its depth map to satisfy the geometric consistency constraints. This indicates that our method correctly learns the object's solid geometry. However, this can degrade the rendered image quality, as evidenced by the blur on the back surface of the bottle in the rendered image in Fig.~\ref{fig:depthloss}-(c). In the Appearance Rendering Path, the bottle correctly becomes transparent again while other objects remain unaffected. This shows that the auxiliary opacity parameter, $\alpha'$, successfully learns to model the opacity of translucent materials.

\noindent\textbf{Depth Prior Loss}: Fig.~\ref{fig:depthloss}-(f) illustrates the impact of the Depth Prior Loss on weakly-featured regions. It is clear that after introducing the depth prior, poorly-observed or weakly-textured areas are recovered more effectively. It is crucial to note that while our loss is applied to the Geometric Regularization Path, the depth map shown in Fig.~\ref{fig:depthloss}-(f) is from the Appearance Rendering Path. This demonstrates how optimizing the geometric path positively influences the appearance Refinement path. Our method successfully leverages the corrective capabilities of the depth prior to improve geometry in challenging regions, all while the decoupling mechanism prevents the precision loss typically associated with depth supervision from affecting the final high-quality output.

\section{Conclusion and Limitation}
\label{sec:conclusion}


This paper addresses the local minima problem in 3DGS training that causes persistent floater artifacts, which we trace back to a gradient vanishing phenomenon. We propose StableGS, a novel method centered on a Dual Opacity GS architecture. This architecture decouples geometry from appearance, providing a robust foundation for applying a synergistic set of geometric constraints-including a self-supervised consistency loss and a globally optimized prior-without harming rendering quality. Our method achieves state-of-the-art performance, successfully eliminating floaters while avoiding the common blur-artifact trade-off and excelling in challenging weakly-textured scenes. We also recognize several limitations and directions for future work. Our use of paired images and auxiliary opacity parameters increases memory and computational overhead. Furthermore, the reliance on sufficient multi-view co-visibility for our depth consistency loss currently limits the method's applicability to few-shot novel view synthesis tasks, which we leave for future investigation.


\bibliography{aaai2026}

\end{document}